\documentclass[11pt]{article}
\usepackage[margin=1in]{geometry}
\usepackage[T1]{fontenc}
\input{glyphtounicode}\pdfgentounicode=1
\usepackage{booktabs,xcolor,amsmath}
\usepackage[hidelinks]{hyperref}
\usepackage{url}
\newcommand{\todo}[1]{}

\title{SpecRead: A Benchmark for Measuring Whether Language Models Understand Hardware Specifications}
\author{Feilian Huang and Yujia Liu\\Independent Researcher}
\date{September 2026}

\begin{document}

\newcommand{\NTotal}{383}
\newcommand{\NTone}{58}
\newcommand{\NTtwo}{58}
\newcommand{\NTthree}{171}
\newcommand{\NTfour}{96}
\newcommand{\NCalls}{465}
\newcommand{\ConsToverall}{129/383}
\newcommand{\ConsAccoverall}{33.7}
\newcommand{\ConsCIoverall}{[0.291, 0.386]}
\newcommand{\ConsTone}{32/58}
\newcommand{\ConsAccTone}{55.2}
\newcommand{\ConsCITone}{[0.425, 0.673]}
\newcommand{\ConsTtwo}{30/58}
\newcommand{\ConsAccTtwo}{51.7}
\newcommand{\ConsCITtwo}{[0.392, 0.641]}
\newcommand{\ConsTthree}{42/171}
\newcommand{\ConsAccTthree}{24.6}
\newcommand{\ConsCITthree}{[0.187, 0.315]}
\newcommand{\ConsTfour}{25/96}
\newcommand{\ConsAccTfour}{26.0}
\newcommand{\ConsCITfour}{[0.183, 0.356]}
\newcommand{\ConsMacroAvg}{39.4}
\newcommand{\ConsOpChangedNumber}{21/53}
\newcommand{\ConsAccOpChangedNumber}{39.6}
\newcommand{\ConsOpCrossrefConflict}{9/30}
\newcommand{\ConsAccOpCrossrefConflict}{30.0}
\newcommand{\ConsOpRuleInversion}{12/51}
\newcommand{\ConsAccOpRuleInversion}{23.5}
\newcommand{\ConsOpDeletedCondition}{0/37}
\newcommand{\ConsAccOpDeletedCondition}{0.0}
\newcommand{\DistrFPR}{21/41}
\newcommand{\DistrAccFPR}{51.2}
\newcommand{\DistrCIFPR}{[0.365, 0.657]}
\newcommand{\CtrlFPR}{41/41}
\newcommand{\CtrlAccFPR}{100.0}
\newcommand{\CtrlCIFPR}{[0.914, 1.000]}
\newcommand{\ConsWithControls}{149/465}
\newcommand{\ConsAccWithControls}{32.0}
\newcommand{\ConsCIWithControls}{[0.280, 0.364]}
\newcommand{\ConsWithGrayOK}{162/383}
\newcommand{\ConsAccWithGrayOK}{42.3}
\newcommand{\ConsCIWithGrayOK}{[0.375, 0.473]}
\newcommand{\DecTP}{95}
\newcommand{\DecFN}{1}
\newcommand{\DecFP}{41}
\newcommand{\DecTN}{0}
\newcommand{\DecBalAcc}{49.5}
\newcommand{\DecMCC}{-0.06}
\newcommand{\DecPrec}{69.9}
\newcommand{\DecRec}{99.0}
\newcommand{\DecFone}{81.9}
\newcommand{\ConsVLTthree}{82/171}
\newcommand{\ConsAccVLTthree}{48.0}
\newcommand{\ConsCIVLTthree}{[0.406, 0.554]}
\newcommand{\ConsVLTfour}{63/96}
\newcommand{\ConsAccVLTfour}{65.6}
\newcommand{\ConsCIVLTfour}{[0.557, 0.744]}
\newcommand{\LayAccTthreeCat}{49.7}
\newcommand{\LayTthreeCat}{85/171}
\newcommand{\LayAccTfourCat}{44.8}
\newcommand{\LayTfourCat}{43/96}
\newcommand{\LayAccTthreeLoc}{78.9}
\newcommand{\LayTthreeLoc}{135/171}
\newcommand{\LayAccTthreeBoth}{40.4}
\newcommand{\LayTthreeBoth}{69/171}
\newcommand{\LayAccTfourLoc}{79.2}
\newcommand{\LayTfourLoc}{76/96}
\newcommand{\LayAccTfourBoth}{36.5}
\newcommand{\LayTfourBoth}{35/96}
\newcommand{\TthreeVerdTP}{171}
\newcommand{\TthreeVerdFN}{0}
\newcommand{\TthreeVerdFP}{21}
\newcommand{\TthreeVerdTN}{20}
\newcommand{\TthreeVerdBalAcc}{74.4}
\newcommand{\TthreeVerdTPR}{100.0}
\newcommand{\TthreeVerdTNR}{48.8}
\newcommand{\ConfTThreeNumNum}{39}
\newcommand{\ConfTThreeNumRule}{3}
\newcommand{\ConfTThreeNumCross}{8}
\newcommand{\ConfTThreeNumMiss}{0}
\newcommand{\ConfTThreeRuleNum}{4}
\newcommand{\ConfTThreeRuleRule}{20}
\newcommand{\ConfTThreeRuleCross}{19}
\newcommand{\ConfTThreeRuleMiss}{2}
\newcommand{\ConfTThreeCrossNum}{9}
\newcommand{\ConfTThreeCrossRule}{4}
\newcommand{\ConfTThreeCrossCross}{24}
\newcommand{\ConfTThreeCrossMiss}{2}
\newcommand{\ConfTThreeMissNum}{3}
\newcommand{\ConfTThreeMissRule}{14}
\newcommand{\ConfTThreeMissCross}{18}
\newcommand{\ConfTThreeMissMiss}{2}
\newcommand{\ConfTFourNumNum}{12}
\newcommand{\ConfTFourNumRule}{26}
\newcommand{\ConfTFourNumCross}{4}
\newcommand{\ConfTFourNumMiss}{7}
\newcommand{\ConfTFourRuleNum}{0}
\newcommand{\ConfTFourRuleRule}{21}
\newcommand{\ConfTFourRuleCross}{2}
\newcommand{\ConfTFourRuleMiss}{2}
\newcommand{\ConfTFourCrossNum}{3}
\newcommand{\ConfTFourCrossRule}{1}
\newcommand{\ConfTFourCrossCross}{1}
\newcommand{\ConfTFourCrossMiss}{0}
\newcommand{\ConfTFourMissNum}{0}
\newcommand{\ConfTFourMissRule}{5}
\newcommand{\ConfTFourMissCross}{2}
\newcommand{\ConfTFourMissMiss}{9}
\newcommand{\BaseAVtThree}{50.0}
\newcommand{\BaseAVtFour}{50.0}
\newcommand{\BaseRandCat}{25.0}
\newcommand{\BaseDiffN}{57/162}
\newcommand{\BaseDiffAcc}{35.2}
\newcommand{\BaseDiffSkip}{9}
\newcommand{\BaseDiffCI}{[0.283, 0.428]}
\newcommand{\TthreeOpChangedNumber}{53}
\newcommand{\TthreeOpRuleInversion}{51}
\newcommand{\TthreeOpCrossrefConflict}{30}
\newcommand{\TthreeOpDeletedCondition}{37}
\newcommand{\TthreeCatNumericMismatch}{50}
\newcommand{\TthreeCatRuleInversion}{45}
\newcommand{\TthreeCatCrossrefConflict}{39}
\newcommand{\TthreeCatMissingCondition}{37}
\newcommand{\TthreeCatDiff}{9}
\newcommand{\TfourWithRule}{96/96}
\newcommand{\PromptVarN}{41}
\newcommand{\PromptVarFP}{41/41}
\newcommand{\PromptVarFPR}{100.0}
\newcommand{\PromptVarCIFPR}{91.4--100.0}
\newcommand{\RuleTone}{3/10}
\newcommand{\RuleDeltaTone}{-0.300}
\newcommand{\RuleTtwo}{4/10}
\newcommand{\RuleDeltaTtwo}{0.000}
\newcommand{\RuleTthree}{7/20}
\newcommand{\RuleDeltaTthree}{-0.100}
\newcommand{\RuleTfour}{0/20}
\newcommand{\RuleDeltaTfour}{-0.150}
\newcommand{\RuleOverall}{14/60}
\newcommand{\RuleDeltaOverall}{-0.133}
\newcommand{\RuleMcNemarB}{10}
\newcommand{\RuleMcNemarC}{2}
\newcommand{\RuleMcNemarP}{0.039}
\newcommand{\AblWithTone}{6/10}
\newcommand{\AblWithAccTone}{60.0}
\newcommand{\AblWithCITone}{[0.313, 0.832]}
\newcommand{\AblWithTtwo}{4/10}
\newcommand{\AblWithAccTtwo}{40.0}
\newcommand{\AblWithCITtwo}{[0.168, 0.687]}
\newcommand{\AblWithTthree}{9/20}
\newcommand{\AblWithAccTthree}{45.0}
\newcommand{\AblWithCITthree}{[0.258, 0.658]}
\newcommand{\AblWithTfour}{3/20}
\newcommand{\AblWithAccTfour}{15.0}
\newcommand{\AblWithCITfour}{[0.052, 0.360]}
\newcommand{\AblWithOverall}{22/60}
\newcommand{\AblWithAccOverall}{36.7}
\newcommand{\AblWithCIOverall}{[0.256, 0.493]}
\newcommand{\AblWithoutTone}{0/10}
\newcommand{\AblWithoutAccTone}{0.0}
\newcommand{\AblWithoutCITone}{[0.000, 0.278]}
\newcommand{\AblWithoutTtwo}{3/10}
\newcommand{\AblWithoutAccTtwo}{30.0}
\newcommand{\AblWithoutCITtwo}{[0.108, 0.603]}
\newcommand{\AblWithoutTthree}{0/20}
\newcommand{\AblWithoutAccTthree}{0.0}
\newcommand{\AblWithoutCITthree}{[0.000, 0.161]}
\newcommand{\AblWithoutTfour}{0/20}
\newcommand{\AblWithoutAccTfour}{0.0}
\newcommand{\AblWithoutCITfour}{[0.000, 0.161]}
\newcommand{\AblWithoutOverall}{3/60}
\newcommand{\AblWithoutAccOverall}{5.0}
\newcommand{\AblWithoutCIOverall}{[0.017, 0.137]}
\newcommand{\AblDeltaTone}{60.0}
\newcommand{\AblDeltaTtwo}{10.0}
\newcommand{\AblDeltaTthree}{45.0}
\newcommand{\AblDeltaTfour}{15.0}
\newcommand{\AblDeltaOverall}{31.7}
\newcommand{\AblWithoutToneTwo}{3/20}
\newcommand{\AblWithoutAccToneTwo}{15.0}
\newcommand{\AblWithoutCIToneTwo}{[0.052, 0.360]}
\newcommand{\LenOverall}{166/383}
\newcommand{\LenAccoverall}{43.3}
\newcommand{\LenCIoverall}{[0.385, 0.483]}
\newcommand{\LenTone}{32/58}
\newcommand{\LenAccTone}{55.2}
\newcommand{\LenCITone}{[0.425, 0.673]}
\newcommand{\LenTtwo}{30/58}
\newcommand{\LenAccTtwo}{51.7}
\newcommand{\LenCITtwo}{[0.392, 0.641]}
\newcommand{\LenTthree}{69/171}
\newcommand{\LenAccTthree}{40.4}
\newcommand{\LenCITthree}{[0.333, 0.478]}
\newcommand{\LenTfour}{35/96}
\newcommand{\LenAccTfour}{36.5}
\newcommand{\LenCITfour}{[0.275, 0.464]}
\newcommand{\LenMacroAvg}{45.9}
\newcommand{\DcLocOnly}{28/37}
\newcommand{\DcLocOnlyAcc}{75.7}

\newcommand{\AdjudN}{93}
\newcommand{\AdjudAgree}{83/93}
\newcommand{\AdjudAgreePct}{89.2}
\newcommand{\AdjudKappa}{0.77}
\newcommand{\AdjudKappaNoLeak}{0.72}
\newcommand{\AdjudNoLeakN}{60}
\newcommand{\AdjudNoLeakAgree}{52/60}
\newcommand{\AdjudNoLeakAgreePct}{86.7}
\newcommand{\AdjudFirstOK}{36/93}
\newcommand{\AdjudFirstAccOK}{38.7}
\newcommand{\AdjudOpusOK}{36/93}
\newcommand{\AdjudOpusAccOK}{38.7}
\newcommand{\AdjudMergedOK}{33/93}
\newcommand{\AdjudMergedAccOK}{35.5}
\newcommand{\AdjudMergedWrong}{60/93}
\newcommand{\AdjudBothWrong}{52}
\newcommand{\AdjudBothOK}{31}
\newcommand{\AdjudFirstWrongOpusOK}{5}
\newcommand{\AdjudFirstOKOpusWrong}{5}
\newcommand{\AdjudDisagree}{10}
\newcommand{\AdjudFirstTranscriptionErr}{2}
\newcommand{\AdjudVOneWins}{5}
\newcommand{\AdjudVTwoWins}{5}

\maketitle

\begin{abstract}
Existing benchmarks for large language models (LLMs) in hardware design evaluate downstream artifacts such as generated RTL, assertions, or testbenches. When a model fails such a benchmark, the failure is ambiguous: it may have misread the specification, or it may have understood the specification and failed to write the code. We present SpecRead, a benchmark that isolates specification comprehension from generation ability. SpecRead v2.2 contains \NTotal\ questions over 10 open-source OpenTitan IP blocks, in four types: exact retrieval, cross-section reasoning, contradiction detection in mutated specifications, and spec--RTL consistency checking, plus 41 distractor controls and 41 consistent-RTL controls. Type-4 items are built from real RTL mutations; we retain only mutations that Icarus Verilog simulation shows to change observable behavior. A with-spec vs.\ without-spec ablation suggests the questions require the excerpt rather than training recall alone, though it does not rule out memorization of the unmutated source text helping a model spot mutations. As an initial characterization with a small model, Ministral-3B scores \ConsAccoverall\% overall (\ConsToverall). On spec contradiction detection (t3), the primary verdict-plus-location measure gives \ConsAccVLTthree\%; the distractor false-positive rate is \DistrAccFPR\%. On spec--RTL consistency (t4), the violation-vs.-compliant decision is effectively degenerate (MCC \DecMCC, balanced accuracy \DecBalAcc\%), with the model flagging a violation on all 41 consistent-RTL controls; location accuracy remains measurable at \LayAccTfourLoc\% (lenient threshold, overlap $\ge 0.25$). Layered scoring shows the model locates contradictions well but scores lower on their fine-grained category, suggesting as a hypothesis that the violation taxonomy is the hard part, though the category labels have not been independently validated. A structured ``rule-table'' prompting intervention on a 60-question sample lowers accuracy on every question type except t2 (tied). SpecRead is automatically scorable by deterministic checks, with gray-zone cases counted wrong under the conservative main scoring; a deterministic lenient-threshold variant is reported as a sensitivity analysis in the appendix. The benchmark is regenerable for type-3 items via mutation injection, and built exclusively from public sources.
\end{abstract}

\section{Introduction}
Large language models are increasingly applied to hardware design: writing RTL, generating assertions, and summarizing specifications. The benchmarks used to evaluate them, including VerilogEval~\cite{verilogeval} and RTLLM~\cite{rtllm}, predominantly measure \emph{generation}: given a specification or natural-language description, produce an artifact and check it by simulation or formal equivalence.

Generation benchmarks conflate two distinct capabilities. Consider a model that writes incorrect RTL for a UART baud-rate divider. Did it misread the specification (a comprehension failure), or did it understand the specification and fail at Verilog syntax, timing discipline, or state-machine encoding (a production failure)? The benchmark cannot tell, and the distinction determines whether to invest in spec-grounded reasoning or in code synthesis.

SpecRead measures comprehension directly. Instead of asking the model to produce hardware, we ask it to \emph{read}: retrieve facts from a specification, reason across sections, detect contradictions planted in mutated specifications, and judge whether an RTL excerpt is consistent with the specification it claims to implement. Every question is scored by deterministic checks, with gray-zone cases counted wrong under the conservative main scoring (see Scoring below). The focus is on types 3 and 4, which have no analogue in generation benchmarks:
\begin{itemize}
\item \textbf{Type 3 (contradiction detection).} We apply controlled mutations (number changes, rule inversions, cross-reference conflicts, deleted conditions) to real specification text and ask the model to locate and classify the contradiction. Distractor items, real unmutated passages that merely look suspicious, control the false-positive rate.
\item \textbf{Type 4 (spec--RTL consistency).} We mutate real RTL, verify by simulation that the mutation changes observable behavior, and ask whether the mutated RTL still matches the specification. The model must connect natural-language rules to Verilog semantics.
\end{itemize}
Our contributions are: (1) the SpecRead benchmark (\NTotal\ questions, 10 IPs, automatically scored); (2) a mutation-injection methodology that makes type-3 items regenerable and therefore more resistant to training-data contamination; (3) type-4 items filtered by simulation to guarantee an observable behavioral change; and (4) an initial single-model characterization on a small open model, with a fine-grained failure taxonomy. We frame the paper as a benchmark release plus this single-model characterization; with one 3B model evaluated, we make no claim that the difficulty profile is discriminative across models or that it transfers to larger or frontier models.

\section{Related Work}
\paragraph{RTL generation benchmarks.} VerilogEval~\cite{verilogeval,verilogeval2} and RTLLM~\cite{rtllm} established the paradigm of evaluating LLMs by the functional correctness of generated Verilog. Follow-up work extended it to assertions (AssertLLM~\cite{assertllm}, AssertLLM2~\cite{assertllm2}), testbenches (AutoBench~\cite{autobench}), and multi-turn repair. These benchmarks measure the end of the pipeline; none isolates whether the model understood its input.

\paragraph{Contamination-resistant evaluation.} Regenerable items, private held-out sets, and dynamic benchmark construction are active concerns across LLM evaluation~\cite{dynabench,livebench}. SpecRead's mutation injection follows this spirit: because type-3 items are generated by scripted mutation of public text, fresh instances can be produced on demand. (Type-4 items require hand-built testbenches and simulation filtering, so they are not regenerable by the same scripted pipeline; see Limitations.)

\paragraph{Indirect evaluations of comprehension.} The closest existing work evaluates comprehension indirectly: generating RTL from a specification and then checking the RTL (round-trip), analyzing characteristic spec-misreading errors, or the reverse direction of RTL-to-spec summarization (RTL-BenchLS~\cite{rtlbenschls}). CVDP~\cite{cvdp} includes an RTL-to/from-specification correspondence category (34 problems), but scores it with BLEU and LLM-based judging rather than deterministic gold labels. The nearest neighbors on each side are VClare~\cite{vclare}, which injects specification defects including contradictions but measures repair outcomes (Verilog pass rate) rather than comprehension, and AssertLLM2~\cite{assertllm2}, whose bug-hunting setting uses mutated RTL but scores generated assertions rather than direct consistency judgments, without simulation-filtering mutations for observable difference. To our knowledge, no existing benchmark directly scores specification comprehension itself: asking the model to detect, locate, and classify contradictions in a specification, or to judge spec--RTL consistency, against gold labels with scoring decoupled from generation ability. Characteristic spec-misreading failures (``Type~II'' errors) are a documented confound in RTL generation~\cite{zhang2025errors}; SpecRead isolates exactly that confound.

\paragraph{AI for EDA.} Domain-adapted models such as ChipNeMo~\cite{chipnemo} and EDA copilots presuppose that models can read hardware documentation faithfully. SpecRead provides a direct measurement of that presupposition. Recent EDA question-answering work (Ask-EDA~\cite{askeda}, ORD-QA~\cite{ordqa}, the EDA Corpus~\cite{edacorpus}) evaluates models on documentation-grounded QA but does not test contradiction detection or spec--RTL consistency.

\paragraph{Natural language inference and contradiction detection.} Type-3 items are a domain-specific form of contradiction detection over long documents. General NLI benchmarks (DocNLI~\cite{docnli}, ContractNLI~\cite{contractnli}) and factual-consistency metrics (SummaC~\cite{summac}) test whether models detect inconsistencies between texts, but on news or contract domains with different linguistic structure than hardware specifications. Long-document QA and contextual NLI (ConTRoL~\cite{control}) share the multi-hop reasoning demand of type-2 items. SpecRead differs in grounding contradictions in executable hardware semantics: type-4 mutations are simulation-verified to change observable behavior, a guarantee no text-only NLI benchmark provides.

\section{Benchmark Design}
\subsection{Sources}
All materials derive from the OpenTitan project~\cite{opentitan} (Apache~2.0) at commit \texttt{80b5b321\ldots} (full hash in the Reproducibility section). We record repository, file path, commit, and license for every excerpt. No proprietary or company-internal specifications are used.

\subsection{Question types}
\paragraph{Type 1: exact retrieval ($n{=}58$).} Answer a factual question whose answer appears verbatim in the excerpt (e.g., a reset value or register address).

\paragraph{Type 2: cross-section reasoning ($n{=}58$).} The answer requires combining statements from two or more sections (e.g., a programming sequence spanning the register list and the theory-of-operation section). Of the 58, 22 are multiple-choice with four options presented in the prompt (an earlier version omitted the option list on 15 of these; all were repaired in v2.1).

\paragraph{Type 3: mutated-spec contradiction detection ($n{=}171$).} We apply one scripted mutation to a real excerpt and ask the model either to identify the contradiction (location and category) or to reply \texttt{NO\_CONTRADICTION}. Mutation operators are \emph{changed number} (e.g., ``9/11/13 rounds'' instead of the documented values), \emph{rule inversion} (polarity flips), \emph{cross-reference conflict} (a table row contradicting prose), and \emph{deleted condition} (a dropped qualifier creating an over-generalization). The corresponding categories are \texttt{NUMERIC\_MISMATCH}, \texttt{RULE\_INVERSION}, \texttt{CROSSREF\_CONFLICT}, and \texttt{MISSING\_CONDITION}. Mutations are applied by exact-match scripted replacement, and the mutated sentence is kept fluent to avoid ``looks edited'' cues. Operator counts: \texttt{changed\_number} \TthreeOpChangedNumber, \texttt{rule\_inversion} \TthreeOpRuleInversion, \texttt{deleted\_condition} \TthreeOpDeletedCondition, \texttt{crossref\_conflict} \TthreeOpCrossrefConflict. The resulting gold categories are \texttt{NUMERIC\_MISMATCH} \TthreeCatNumericMismatch, \texttt{RULE\_INVERSION} \TthreeCatRuleInversion, \texttt{CROSSREF\_CONFLICT} \TthreeCatCrossrefConflict, \texttt{MISSING\_CONDITION} \TthreeCatMissingCondition (\TthreeCatDiff\ items' gold category differs from the generating operator's default mapping).

\paragraph{Type 4: spec--RTL consistency ($n{=}\NTfour$).} We mutate real RTL (single-line substitutions or small deletions), simulate the original and mutated modules under Icarus Verilog~\cite{iverilog} with a purpose-built testbench, and retain only mutations whose waveforms differ. Candidate RTL mutations were proposed by hand for each IP and simulated under purpose-built testbenches; only mutations with \texttt{RESULT: DIFFER} were retained, leaving \NTfour. No single candidate/discard ledger was kept; per-IP review notes document 26 discarded candidates, dropped for behavioral equivalence under the testbench, Icarus Verilog quirks, or lack of a spec witness. A behavioral difference from the original RTL does not by itself imply a specification violation, since the changed behavior could fall in an unspecified region; at item-authoring time the first author therefore wrote, for each of the \NTfour\ retained items, a gold answer citing the specific spec clause the mutation violates (a \texttt{rule} field plus an explanation linking the mutated line to that clause, present in \NTfour/\NTfour\ items). The model receives the specification excerpt and the mutated RTL excerpt and must judge consistency, citing the location and one of the four categories. Category mapping for RTL mutations: \texttt{NUMERIC\_MISMATCH} (50 items) for mutations changing a numeric constant, width, or count contradicting a spec-stated value; \texttt{RULE\_INVERSION} (25) for mutations inverting a logical condition or polarity (e.g., active-high vs.\ active-low); \texttt{MISSING\_CONDITION} (16) for mutations deleting a guard, check, or reset the spec requires; \texttt{CROSSREF\_CONFLICT} (5) for mutations whose violation is visible only by cross-referencing two spec clauses (e.g., a bark/bite threshold swap where each register is defined separately). The \texttt{NUMERIC\_MISMATCH}/\texttt{RULE\_INVERSION} boundary is genuinely ambiguous for mutations like a numeric change that inverts a comparison; see Failure Analysis.

\subsection{Distractor controls}
The 41 distractor items are real, unmutated excerpts that look suspicious but are self-consistent: reset-value subtleties, level- vs.\ event-triggered interrupt semantics, reserved register fields, \texttt{x} reset values, and read-to-clear behavior. They are interleaved with the main set, so a model that flags anything unusual as a contradiction accumulates false positives. Of an initial 42 distractors, one (\texttt{uart-t3d-003}) was dropped as an exact duplicate prompt of \texttt{uart-t3d-001}. Distractors are presented in the type-3 format.

We also built 41 consistent-RTL controls for type~4: real, unmutated RTL excerpts paired with their specification passages, presented in the type-4 format, for which the correct verdict is \texttt{NO\_CONTRADICTION}. Each control was individually audited by the first author for fairness (the excerpt must contain enough context to verify consistency); 24 excerpts were widened and 3 unfair controls replaced. These controls measure the false-positive rate on the violation-vs.-compliant decision specifically. (This fairness audit of controls is distinct from review of the \NTfour\ t4 gold labels, which no second annotator reviewed; see Limitations.)

\subsection{Scoring}
The main results use conservative automatic scoring: deterministic checks only, with every gray-zone case counted wrong. Type~1 uses strict normalized exact match; type~2 uses exact match or multiple-choice letter match. For types~3 and~4 the \emph{primary} measure is verdict plus location: did the model find the contradiction and cite the right place? The category label is a \emph{secondary}, finer-grained measure, since the four violation categories are not perfectly mutually exclusive (e.g., a numeric change inside a cross-referenced table). A cited location is checked by distinctive-token overlap against the gold location: overlap $\ge 0.45$ counts as correct, $< 0.25$ as wrong, and the gray zone in between is counted wrong under the conservative scoring, with no human or LLM judgment involved. As a precision guard, citations quoting more than five times the gold location's distinctive tokens fall into the gray zone rather than auto-scoring, so the recall-style overlap check cannot be gamed with overly broad quotes. Full credit on an item requires the primary verdict-plus-location plus the correct category. The conservative scores are fully reproducible by running \texttt{score\_main.py --conservative} and \texttt{score\_layered.py --conservative} in the released repository.

Separately, a deterministic lenient-threshold variant scores t3/t4 items with the location overlap bar lowered from 0.45 to 0.25 (the ``not wrong'' boundary); t1/t2 are unchanged, as exact match has no gray zone. It is reported only as a sensitivity analysis in Appendix~A, showing how much the headline figures move under a more forgiving location rule. Distractors are correct if and only if the verdict is ``consistent''. Prompts are built once per question (verified by prompt SHA-256). We report 95\% Wilson score intervals~\cite{wilson} throughout.

To check the gray-zone adjudication rubric, \AdjudN\ gray-zone t3/t4 responses from the Ministral-3B run underwent an LLM-assisted consistency check: the first annotator was the first author's AI research assistant (Muse Spark by Meta), and the second was Claude Opus~5.5, instructed to judge each item on the rubric alone (same planted location plus exact category match). The second annotation was designed as blinded to the first round's verdicts; on two batches the second annotator was inadvertently shown the first annotator's comparison notes, and confirmed disregarding them and judging from the items alone. Because both annotators are language models, we describe this as a consistency check rather than independent validation. Raw agreement was \AdjudAgree\ (\AdjudAgreePct\%, Cohen's $\kappa=\AdjudKappa$). Excluding the two affected batches, agreement was \AdjudNoLeakAgree\ (\AdjudNoLeakAgreePct\%, $\kappa=\AdjudKappaNoLeak$), so the headline figure is not driven by the leak. All \AdjudDisagree\ disagreements were re-examined against the raw model responses: \AdjudVOneWins\ resolved in favor of the first annotator's reading and \AdjudVTwoWins\ in favor of the second's, including \AdjudFirstTranscriptionErr\ where the first round's notes contained transcription errors, yielding merged labels of \AdjudMergedOK\ correct and \AdjudMergedWrong\ wrong (per-item provenance in \texttt{build/eval/merged\_gray\_labels\_v2.json}). Disagreement concentrated on location-precision standards for partial responses, indicating the rubric is stable except at the boundary cases it was designed to probe. The 93 items comprise all t3/t4 responses where the automatic location check fell into the gray zone (overlap 0.25--0.45) or triggered the precision guard, i.e., every case the conservative scorer counts wrong for location uncertainty; this is a superset of the lenient-minus-strict difference, which counts only gray-zone items whose category was also correct. Two scoring details: (i) the precision guard (cited tokens $>$5$\times$ gold tokens $\to$ wrong) applies only under conservative scoring; under the lenient 0.25 threshold, a precision-guard item counts correct if its overlap meets the bar. (ii) Conservative t4 verdict+location uses the same token-overlap check as t3 (threshold 0.45); the mutation-aware check (cited code must contain the mutated snippet's distinctive tokens) is used only in the layered/lenient scoring, where the threshold is 0.25.

\subsection{Anti-contamination check}
A with-spec vs.\ without-spec ablation on a 60-question sample indicates that the benchmark measures reading rather than recall: without the excerpt, accuracy on the informative t1/t2 subset is \AblWithoutToneTwo, and the three correct answers trace to two leaky type-2 questions (flagged, one since rewritten) and one multiple-choice guess (Section~\ref{sec:ablation}).

\section{Dataset Statistics}
Dataset v2.1 (frozen 2026-09-26) reflected a full audit pass: 41 stale type-4 excerpts were repaired against the actual mutated RTL; all \NTfour\ type-4 mutations were re-simulated and re-confirmed to differ; 10 legacy UART type-4 labels were converted to the canonical four-category scheme; and an exhaustive deduplication audit found exactly one true duplicate pair (an earlier internal report of seven pairs did not reproduce; the other six were distinct mutations). A later gold audit (v2.2, frozen 2026-09-27) found one further functionally identical AES pair and deprecated one item; see Reproducibility for the version history.

\begin{table}[h]\centering
\begin{tabular}{lr}\toprule
IP block & questions\\\midrule
i2c & 55\\ spi\_host & 55\\ aes & 50\\ spi\_device & 50\\ kmac & 45\\ hmac & 35\\ uart & 30\\ aon\_timer & 25\\ rv\_timer & 22\\ pattgen & 18\\\midrule
Total & \NTotal\\\bottomrule
\end{tabular}
\caption{SpecRead composition: \NTotal\ questions across 10 OpenTitan IP blocks (t1: 58, t2: 58, t3: 171, t4: \NTfour), plus 41 distractor controls and 41 consistent-RTL controls.}
\end{table}

\section{Experiments}
\subsection{Main evaluation}
We evaluated Ministral-3B (\texttt{ministral-3b-latest}, the Mistral API alias as served on 2026-09-25/26; the API exposes no static version string, so we log prompts, timestamps, and the alias for reproducibility; temperature 0; one question per call) on all \NTotal\ questions plus 41 distractors plus 41 t4 controls (\NCalls\ calls, 0 errors). Table~\ref{tab:main} reports results. The calls ran in four batches (UTC): 2026-09-25 21:xx, t1 (58), 43 unrepaired t2, t3 (171), and distractors (41); 2026-09-26 01:xx, the 15 repaired t2 items whose prompts gained the previously omitted option lists; 2026-09-26 02:xx, all \NTfour\ t4 items and 41 t4 controls with the v2.1 prompt template; 2026-09-26 03:xx, all 171 t3 items re-run with the published template (see below). A post-freeze audit then found the 09-25 t3 calls used a prompt template differing by 36 characters from the published template (a verdict-commit wording tweak; the ablation with-spec t3 records already used the published template), so all 171 t3 items were re-run on 2026-09-26 with the published template, replacing the stale records in place. Per-record timestamps and prompt SHA-256 hashes are logged in the release, making the mixed timeline auditable; after the re-run, all records are prompt-identical with the published templates.

\begin{table}[h]\centering
\begin{tabular}{lrrrrc}\toprule
Type & $n$ & verdict+loc.\ correct & verdict+loc.\ acc. & full-credit acc. & 95\% CI (v+l)\\\midrule
t1 (exact retrieval) & \NTone & 32 & \ConsAccTone\% & \ConsAccTone\% & \ConsCITone\\
t2 (cross-section reasoning) & \NTtwo & 30 & \ConsAccTtwo\% & \ConsAccTtwo\% & \ConsCITtwo\\
t3 (spec contradiction) & \NTthree & \ConsVLTthree & \ConsAccVLTthree\% & \ConsAccTthree\% & \ConsCIVLTthree\\
t4 (spec--RTL consistency) & \NTfour & \ConsVLTfour & \ConsAccVLTfour\% & \ConsAccTfour\% & \ConsCIVLTfour\\\midrule
Overall & \NTotal & --- & --- & \ConsAccoverall\% (\ConsToverall) & \ConsCIoverall\\
Macro avg.\ (mean of 4 types)\textsuperscript{a} & & & & \ConsMacroAvg\% & \\\midrule
Distractor false-positive rate & 41 & 21 FP & \DistrAccFPR\% & & \DistrCIFPR\\
t4 control false-positive rate & 41 & \CtrlFPR & \CtrlAccFPR\% & & \CtrlCIFPR\\\midrule
\multicolumn{6}{l}{Model-free baselines (deterministic): always-violation verdict balanced acc.\ t3 \BaseAVtThree\%, t4 \BaseAVtFour\%;}\\
\multicolumn{6}{l}{random category \BaseRandCat\%; majority-class category t3 29.2\% (\TthreeCatNumericMismatch/\NTthree), t4 52.1\% (50/\NTfour);}\\
\multicolumn{6}{l}{t3 diff-against-upstream location \BaseDiffAcc\% (\BaseDiffN; 95\% CI \BaseDiffCI).}\\
\multicolumn{6}{l}{\textsuperscript{a}Macro average uses t3/t4 \emph{full-credit} scores (\ConsAccTthree\%, \ConsAccTfour\%), not the verdict-plus-location column.}\\
\multicolumn{6}{l}{For t3/t4 the primary headline is verdict-plus-location; full credit additionally requires the correct category.}\\
\multicolumn{6}{l}{For t1/t2 full credit equals the reported accuracy (exact match, no layered scoring).}\\\bottomrule
\end{tabular}
\caption{Main results, Ministral-3B (dataset v2.2), conservative scoring: gray-zone cases counted wrong. ``Overall'' requires verdict, location, and category on t3/t4. Baselines are computed by script with no model calls (see text).}
\label{tab:main}
\end{table}

Retrieval is the strongest type on full credit at \ConsAccTone\%, and cross-section reasoning reaches \ConsAccTtwo\% after the multiple-choice repair. (The t4 verdict-plus-location figure of \ConsAccVLTfour\% is higher but uses a weaker criterion; on full credit t1 remains the strongest.) On the two contradiction-focused types, which have no generation-benchmark analogue, the primary verdict-plus-location measure gives \ConsAccVLTthree\% for spec contradiction (t3, \ConsVLTthree) and \ConsAccVLTfour\% for spec--RTL consistency (t4, \ConsVLTfour). Full credit (verdict, location, and category) is lower at \ConsAccTthree\% and \ConsAccTfour\%, reflecting the category-classification gap discussed below. For context, the deterministic baselines are \BaseAVtThree\% verdict balanced accuracy for an always-violation predictor (t3 and t4), \BaseRandCat\% expected category accuracy for random labels, and \BaseDiffAcc\% t3 location accuracy for a diff-against-upstream string comparison (\BaseDiffN\ on 162 items; 95\% CI \BaseDiffCI; \BaseDiffSkip\ items skipped where direct mutation reversal was impossible, vs.\ the model's \LayTthreeLoc\ on all 171), which bounds how much of the location sub-task is solvable by memorized-source comparison alone. The diff baseline is modest because the gold location cites the contradiction \emph{pair} (the mutated sentence plus the statement it conflicts with), while diffing recovers only the mutated sentence itself; a perfect diff thus earns only partial overlap with the gold location. Both use the conservative location rule (overlap $\ge 0.45$ with precision guard).

The \DistrAccFPR\% distractor false-positive rate is a finding in its own right. The model is easily misled by suspicious-looking but self-consistent passages; it over-flags rather than under-flags. On the t3 verdict-level decision (171 mutated items, gold: violation; 41 distractors, gold: consistent), the model flags a violation on \TthreeVerdTP/171 mutations and \TthreeVerdFP/41 distractors: TPR \TthreeVerdTPR\%, TNR \TthreeVerdTNR\%, balanced accuracy \TthreeVerdBalAcc\%. This is the one non-degenerate verdict signal in the paper: unlike t4, the model does produce the compliant verdict on self-consistent spec passages about half the time.

The t4 consistent-RTL controls sharpen this: on 41 unmutated RTL excerpts where the correct answer is \texttt{NO\_CONTRADICTION}, the model flagged a violation every single time (\CtrlFPR\ false positives), never once producing the compliant verdict even though the prompt explicitly offered it. Combining the \NTfour\ mutations (gold: violation) with the 41 controls (gold: compliant) into one violation-vs.-compliant decision task, the model predicts ``violation'' on \DecTP/\NTfour\ mutations and \DecFP/41 controls: balanced accuracy \DecBalAcc\%, precision \DecPrec\%, recall \DecRec\%, F1 \DecFone\%, MCC \DecMCC. The near-zero MCC confirms the decision carries almost no signal; for this small model it is effectively degenerate toward ``violation.'' A headline figure that accounts for the controls: counting the \ConsToverall\ conservative-correct questions plus 20 distractor true negatives and 0 t4-control true negatives over all \NCalls\ calls gives \ConsWithControls\ (\ConsAccWithControls\%, 95\% CI \ConsCIWithControls), a control-adjusted overall that penalizes the over-flagging the per-type accuracies miss.

\paragraph{Prompt-sensitivity check.} The 100\% false-positive rate on t4 controls could in principle be an artifact of the prompt's verdict-commit instruction (``You must commit to a verdict''). To test this, we re-ran all \PromptVarN\ t4 controls with a variant prompt whose instruction instead begins ``The RTL excerpt may be fully compliant with the specification'' (everything else byte-identical; same model, temperature 0, deterministic item order). The model still flagged a violation on \PromptVarFP\ controls (\PromptVarFPR\%, 95\% Wilson CI \PromptVarCIFPR). The over-flagging bias is robust to this prompt wording; it is not explained by the verdict-commit instruction alone. (Raw responses and prompt hashes in \texttt{build/eval/responses\_controls\_mistral\_promptvar.jsonl}.)

A 30-question pilot on OpenTitan UART preceded the full build as a difficulty check, run on \texttt{gemini-flash-lite-latest} (16/30) and \texttt{ministral-3b-latest} (14/30). Both pilot models scored below the 60\% bar that would have triggered hardening, and missed most planted contradictions rather than over-flagging, which motivated the full build. (Pilot records are in \texttt{pilot/model\_testing/} in the repository.) The pilot under-flagging contrasts with the main model's over-flagging on distractors and t4 controls, a difference we attribute to the full build's harder distractor set and the t4 control format rather than to a model change. Evaluation of larger models on the identical \NCalls\ prompts is left to a future revision; all numbers in this paper are for Ministral-3B, and we do not claim that the difficulty profile transfers to frontier models.

\subsection{Contamination ablation}\label{sec:ablation}
Table~\ref{tab:ablation} compares with-spec and without-spec accuracy on the 60-question sample. The sample is one item per IP for t1/t2 and two per IP for t3/t4, taking the lowest-numbered IDs per IP: deterministic and stratified by IP, not a uniform random draw. The 60-question sample was scored with the same deterministic conservative scorer as the main results (gray-zone cases counted wrong; the first author's AI research assistant spot-checked the gray-zone items and confirmed the automatic verdicts without changes). The with-spec sample scores \AblWithAccTone\% (t1), \AblWithAccTtwo\% (t2), \AblWithAccTthree\% (t3), \AblWithAccTfour\% (t4), and \AblWithAccOverall\% overall, against conservative full-set scores of \ConsAccTone\%, \ConsAccTtwo\%, \ConsAccTthree\% (full credit; \ConsAccVLTthree\% verdict-plus-location), \ConsAccTfour\% (full credit; \ConsAccVLTfour\% verdict-plus-location), and \ConsAccoverall\%. The ablation's purpose, detecting contamination rather than estimating absolute difficulty, is unaffected: the without-spec column is what carries the claim.

\begin{table}[h]\centering
\begin{tabular}{lccc}\toprule
Type & with spec & without spec & $\Delta$\\\midrule
t1 ($n{=}10$) & \AblWithTone\ = \AblWithAccTone\% & \AblWithoutTone\ = \AblWithoutAccTone\% & +\AblDeltaTone\ pp\\
t2 ($n{=}10$) & \AblWithTtwo\ = \AblWithAccTtwo\% & \AblWithoutTtwo\ = \AblWithoutAccTtwo\% & +\AblDeltaTtwo\ pp\\
t3 ($n{=}20$) & \AblWithTthree\ = \AblWithAccTthree\% & \AblWithoutTthree\ = \AblWithoutAccTthree\% & +\AblDeltaTthree\ pp\\
t4 ($n{=}20$) & \AblWithTfour\ = \AblWithAccTfour\% & \AblWithoutTfour\ = \AblWithoutAccTfour\% & +\AblDeltaTfour\ pp\\\midrule
Overall & \AblWithOverall\ = \AblWithAccOverall\% & \AblWithoutOverall\ = \AblWithoutAccOverall\% & +\AblDeltaOverall\ pp\\\bottomrule
\end{tabular}
\caption{Contamination ablation, Ministral-3B (dataset v2.2, fixed prompts; sample scored with the deterministic conservative scorer, gray-zone cases counted wrong). 95\% Wilson intervals for the overall row: with spec \AblWithCIOverall, without spec \AblWithoutCIOverall. The without-spec prompts contain the question only, no excerpt and no RTL.}
\label{tab:ablation}
\end{table}

The contamination conclusion from this ablation is restricted to t1 and t2: without the excerpt, t1 accuracy drops to \AblWithoutAccTone\% and t2 to \AblWithoutAccTtwo\%, showing these questions require reading the provided text rather than recalling training data. The small t2 delta is expected: 3 of the 10 sampled t2 questions are answerable or guessable without the excerpt, which is exactly what the ablation is designed to detect. For t3 and t4, the without-spec condition is uninformative by construction: the without-spec prompt contains the question only (no excerpt and, for t4, no RTL), so locating a planted contradiction or judging spec--RTL consistency is impossible without the material the question is about. We draw no contamination conclusion for t3/t4 from this ablation. Note further that even for t3, the ablation does not rule out that memorized knowledge of the \emph{unmutated} OpenTitan text helps a model notice a mutation; contamination resistance for the mutated types rests on regenerability (fresh mutations on demand) rather than on this ablation alone.

\subsection{Rule-table prompting (negative result)}
A two-step prompting intervention first extracts the excerpt into a numbered rule table and then answers using only the table. On the same 60-question sample it lowers accuracy on every type except t2 (tied): t1 \RuleDeltaTone\ (\AblWithTone\ to \RuleTone), t2 \RuleDeltaTtwo\ (\AblWithTtwo\ to \RuleTtwo), t3 \RuleDeltaTthree\ (\AblWithTthree\ to \RuleTthree), t4 \RuleDeltaTfour\ (\AblWithTfour\ to \RuleTfour), overall \RuleDeltaOverall\ (\AblWithOverall\ to \RuleOverall). A McNemar exact test on the 60 paired outcomes gives $p=\RuleMcNemarP$ (\RuleMcNemarB\ discordant pairs favoring the standard prompt, \RuleMcNemarC\ favoring rule-table), so the degradation is statistically significant at $\alpha=0.05$. Step-1 extraction is faithful (spot-checked tables preserve the mutated numbers); the loss occurs in step 2 through recombination failures, hedging instead of committing to a verdict, and paraphrase drift penalized by strict scoring. For a 3B model, structured extraction does not help specification reading; it hurts.

\section{Failure Analysis}
Layered scoring separates \emph{verdict/location} (did the model find the contradiction?) from \emph{category} (did it label the violation type correctly?). On t3, the model cites the correct location on \LayTthreeLoc\ items (\LayAccTthreeLoc\%) but the correct category on only \LayTthreeCat\ (\LayAccTthreeCat\%); both correct on \LayTthreeBoth\ (\LayAccTthreeBoth\%). On t4, using a mutation-aware location check (the cited code must contain the mutated snippet's distinctive tokens), location accuracy is \LayTfourLoc\ (\LayAccTfourLoc\%), category \LayTfourCat\ (\LayAccTfourCat\%, below the majority-class baseline of 52.1\% from always predicting \texttt{NUMERIC\_MISMATCH}), both \LayTfourBoth\ (\LayAccTfourBoth\%). The t4 check differs from t3's token-overlap rule because t4 responses cite RTL code: a generic overlap could match surrounding unmutated lines, so the check specifically requires the mutated snippet's tokens, ensuring the model found the actual mutation rather than nearby context. The layered location bar is the lenient ``not wrong'' boundary (overlap $\ge 0.25$), so the ``both'' figures exceed the conservative strict scores (\ConsTthree, \ConsTfour), which count every gray-zone case wrong: the gap between ``both'' and strict is exactly the gray zone. The pattern is the same on both types: the model usually finds the discrepancy but misclassifies it. We state this as a hypothesis, not a firm conclusion: the fine-grained violation taxonomy \emph{may} be the hard part for this model, as opposed to spotting that something is wrong. The hypothesis is confounded with label ambiguity: the four category labels have not been independently validated, and for RTL mutations \texttt{NUMERIC\_MISMATCH} and \texttt{RULE\_INVERSION} often overlap (e.g., a numeric change that inverts a comparison), so part of the ``misclassification'' may reflect genuinely ambiguous gold labels rather than model failure. Category confusion matrices (Tables~\ref{tab:conf3}, \ref{tab:conf4}) show the dominant errors: on t3, the model over-predicts \texttt{CROSSREF\_CONFLICT}, mislabeling \ConfTThreeRuleCross\ \texttt{RULE\_INVERSION} and \ConfTThreeMissCross\ \texttt{MISSING\_CONDITION} items as cross-reference conflicts. On t4 the dominant error is different: \ConfTFourNumRule\ \texttt{NUMERIC\_MISMATCH} items mislabeled as \texttt{RULE\_INVERSION} (vs.\ only \ConfTThreeNumRule\ such errors on t3).

\begin{table}[h]\centering
\begin{tabular}{lrrrr}\toprule
gold $\backslash$ pred & CROSSREF & MISSING & NUMERIC & RULE\_INV\\\midrule
CROSSREF\_CONFLICT & \ConfTThreeCrossCross & \ConfTThreeCrossMiss & \ConfTThreeCrossNum & \ConfTThreeCrossRule\\
MISSING\_CONDITION & \ConfTThreeMissCross & \ConfTThreeMissMiss & \ConfTThreeMissNum & \ConfTThreeMissRule\\
NUMERIC\_MISMATCH & \ConfTThreeNumCross & \ConfTThreeNumMiss & \ConfTThreeNumNum & \ConfTThreeNumRule\\
RULE\_INVERSION & \ConfTThreeRuleCross & \ConfTThreeRuleMiss & \ConfTThreeRuleNum & \ConfTThreeRuleRule\\\bottomrule
\end{tabular}
\caption{t3 category confusion matrix, Ministral-3B (rows: gold, columns: predicted).}
\label{tab:conf3}
\end{table}

\begin{table}[h]\centering
\begin{tabular}{lrrrr}\toprule
gold $\backslash$ pred & CROSSREF & MISSING & NUMERIC & RULE\_INV\\\midrule
CROSSREF\_CONFLICT & \ConfTFourCrossCross & \ConfTFourCrossMiss & \ConfTFourCrossNum & \ConfTFourCrossRule\\
MISSING\_CONDITION & \ConfTFourMissCross & \ConfTFourMissMiss & \ConfTFourMissNum & \ConfTFourMissRule\\
NUMERIC\_MISMATCH & \ConfTFourNumCross & \ConfTFourNumMiss & \ConfTFourNumNum & \ConfTFourNumRule\\
RULE\_INVERSION & \ConfTFourRuleCross & \ConfTFourRuleMiss & \ConfTFourRuleNum & \ConfTFourRuleRule\\\bottomrule
\end{tabular}
\caption{t4 category confusion matrix, Ministral-3B (rows: gold, columns: predicted; full-credit scoring).\protect\footnotemark}
\label{tab:conf4}
\end{table}
\footnotetext{Gold \texttt{NUMERIC\_MISMATCH} totals 50 items: one (\texttt{spi\_host-t4-005}) is omitted from this table because the model response contained no parseable category label, so the displayed predictions sum to 49.}

We observe the following characteristic failure modes for Ministral-3B.
\begin{itemize}
\item \textbf{Right location, wrong category (t3/t4).} The dominant mode on both types, confirmed by the layered scores above: the model finds the mutated line or conflicting statements but misclassifies them (e.g., cites the correct reset-value mutation yet labels it \texttt{RULE\_INVERSION} instead of \texttt{NUMERIC\_MISMATCH}).
\item \textbf{Mutation spotted, conflict not identified (t3).} The model quotes the mutated sentence but pairs it with an unrelated statement instead of the one it actually conflicts with.
\item \textbf{Hedging instead of committing (t3, rule-table condition).} Under the rule-table prompt, the model describes conditions under which a contradiction would exist rather than giving a verdict; this failure mode is specific to the rule-table intervention. In the main t3 evaluation the model commits to a verdict on all 171 items (171/171 violation judgments); hedge responses that lack a parseable verdict are scored wrong by the automatic parser.
\item \textbf{Wrong contradiction (t4).} The model finds a discrepancy other than the planted one (e.g., flags a word-size difference while the gold is a FIFO-ordering rule).
\item \textbf{Over-flagging (distractors and t4 controls).} \DistrFPR\ false positives on self-consistent passages, and \CtrlFPR\ on consistent RTL.
\end{itemize}

Two concrete cases illustrate the failure modes. On \texttt{kmac-t3-019} (gold: \texttt{CROSSREF\_CONFLICT} for a mutated \texttt{ERR\_CODE} link), the model correctly labeled the category but discussed the \texttt{kmac\_done} interrupt versus \texttt{STATUS.squeeze} polling, a real tension in the excerpt but not the planted mutation: it found \emph{a} conflict, not \emph{the} conflict. On \texttt{aes-t3-001} (gold: \texttt{NUMERIC\_MISMATCH}, 9/11/12 rounds versus 12/14/16 cycles), the model quoted the mutated ``9/11/12 rounds'' figure but paired it with an unrelated throughput sentence instead of the conflicting cycle counts, so the contradiction was never isolated. Both cases show the model engaging with the right region of the spec while missing the precise planted inconsistency.

A natural next step is to test whether SpecRead scores predict downstream generation failures: do models that misread specifications here also commit more specification-misreading errors on VerilogEval-style generation tasks? The failure modes above (right region, wrong precise conflict) suggest the errors would transfer, but a quantitative correlation study is left to future work.
By mutation operator on t3, numeric edits are the easiest to catch and dropped conditions the hardest (Table~\ref{tab:operators}): the model notices altered numbers but never gets a \texttt{deleted\_condition} item fully right (\ConsOpDeletedCondition\ strict), the subtlest operator. A location-only re-analysis of the 37 \texttt{deleted\_condition} items (counting a response as finding the conflict when its cited location overlaps the gold location above the automatic-wrong threshold, ignoring category) rises from \ConsOpDeletedCondition\ strict to \DcLocOnly\ (\DcLocOnlyAcc\%), confirming that even on the hardest operator the model usually sees the conflict but cannot name its type.
\begin{table}[h]\centering
\begin{tabular}{lrrc}\toprule
Operator & correct & $n$ & acc.\\\midrule
\texttt{changed\_number} & 21 & 53 & \ConsAccOpChangedNumber\\
\texttt{crossref\_conflict} & 9 & 30 & \ConsAccOpCrossrefConflict\\
\texttt{rule\_inversion} & 12 & 51 & \ConsAccOpRuleInversion\\
\texttt{deleted\_condition} & 0 & 37 & \ConsAccOpDeletedCondition\\\bottomrule
\end{tabular}
\caption{Type-3 accuracy by mutation operator, Ministral-3B (conservative scoring).}
\label{tab:operators}
\end{table} Strict exact-match scoring penalizes paraphrase-correct t1/t2 answers, but a lenient variant (the pipeline's \texttt{score\_exact\_lenient}, which strips code formatting) flips only 4 of the 94 non-multiple-choice t1/t2 judgments on the full set, so the headline numbers stand.

\section{Limitations}
(1) Only one small open model (Ministral-3B) is evaluated; conclusions about difficulty may not transfer to larger or frontier models, which we did not evaluate (resource constraint). (2) All sources come from a single project (OpenTitan); cross-project generality is untested. (3) Specifications are English prose plus register tables; other documentation styles (e.g., assertions as specification) are not covered. (4) Type-4 items use small, simulation-observable mutations; deep microarchitectural changes are out of scope, and a simulation difference from the original RTL is a proxy for, not a proof of, a specification violation. (5) Strict exact-match scoring conflates retrieval with formatting discipline for small models. (6) The main results use conservative scoring (gray-zone cases counted wrong). A deterministic lenient-threshold variant (location overlap $\ge 0.25$) appears only in the appendix sensitivity analysis; no second annotator reviewed the \NTfour\ t4 gold labels, and the category labels have not been independently validated (see Failure Analysis). The gray-zone adjudication rubric itself was checked by LLM-assisted double annotation ($\kappa=\AdjudKappa$ on \AdjudN\ items, $\kappa=\AdjudKappaNoLeak$ excluding two batches with inadvertent exposure; see Scoring). (7) The t4 control false-positive rate (\CtrlFPR) rests on 41 items; the controls were fairness-audited but a larger control set would tighten the interval. Combining the \NTfour\ mutations with the 41 controls, the violation-vs.-compliant decision has balanced accuracy \DecBalAcc\% and MCC \DecMCC\ (see Results): for this small model the decision is effectively degenerate. (8) Regenerability via mutation injection applies to type-3 items; type-4 items require hand-built testbenches and simulation filtering, so they are not regenerable by the same scripted pipeline. No hidden held-out test set is currently maintained. (9) No human baseline is reported; the benchmark measures model behavior, not human difficulty. (10) Post-hoc gold audit (2026-09-27): a review by Claude Opus 5.5 of the nine flagged gold labels found two items indefensible (one t4 item whose gold rule does not govern the mutated behavior with no supporting specification text; one AES item functionally identical to another), both removed from the dataset (v2.2, 383 questions); seven further items received excerpt expansions or clarified gold explanations without changing their gold verdicts.

\section{Conclusion}
SpecRead separates understanding a hardware specification from producing hardware artifacts. As a benchmark release with an initial single-model characterization, it shows that on a small open model there is a clear gap: \ConsAccoverall\% overall (\ConsToverall; macro average \ConsMacroAvg\%), \ConsAccVLTthree--\ConsAccVLTfour\% verdict-plus-location on the two contradiction-focused types that have no generation-benchmark analogue, a \DistrAccFPR\% false-positive rate on self-consistent passages, and a 100\% false-positive rate on consistent-RTL controls. Layered scoring shows the model usually locates contradictions (\LayAccTthreeLoc--\LayAccTfourLoc\%) but scores lower on their category (\LayAccTfourCat--\LayAccTthreeCat\%), suggesting as a hypothesis that the fine-grained violation taxonomy may be the hard part, subject to the label-ambiguity caveat above. We make no claim that this difficulty profile transfers to larger or frontier models. The benchmark is automatically scorable under conservative scoring, regenerable for type-3 items by mutation injection, simulation-grounded, and built entirely from public sources. The dataset, simulation harnesses, and scoring pipeline are publicly released at \url{https://github.com/EDGAhab/specread} so the community can measure, and improve, models' ability to read the documents hardware is built from.

\section*{Reproducibility}
Dataset v2.2 frozen 2026-09-27: 383 questions, 41 distractors, 41 t4 controls; sources pinned to OpenTitan commit \texttt{80b5b3213453156fe62683962664d1afd9695ec4} (Apache~2.0); all type-4 mutations re-simulated under Icarus Verilog and confirmed to differ; temperature 0; per-question prompts verified by SHA-256. Dataset versions: v1 (pilot, 30 questions, frozen 2026-09-24), v2 (full build, 385 questions, frozen 2026-09-25), v2.1 (385 questions, frozen 2026-09-26; t2 multiple-choice repair, t4 prompt fix, 41 t4 controls added), v2.2 (383 questions, frozen 2026-09-27; gold audit: 2 items deprecated, 7 items excerpt/explanation fixes, of which 4 t3/t4 items with changed excerpts were re-run with the published template). Difficulty labels (easy/hard) are recorded per item in the release. Code and data: publicly released at \url{https://github.com/EDGAhab/specread}.

\appendix
\section{Lenient-threshold scoring: sensitivity analysis}\label{app:lenient}
The main results count every gray-zone case wrong (location overlap below 0.45). As a deterministic sensitivity analysis, Table~\ref{tab:len} reports the scores when the location overlap bar is lowered to 0.25 (the ``not wrong'' boundary used in the layered analysis); t1/t2 are unchanged, as exact match has no gray zone. This variant is fully scripted with no human or model judgment, and shows how much the headline figures move under a more forgiving location rule.

\begin{table}[h]\centering
\begin{tabular}{lrrrc}\toprule
Type & $n$ & correct & acc.\ (lenient) & 95\% CI\\\midrule
t1 (exact retrieval) & \NTone & 32 & \LenAccTone\% & \LenCITone\\
t2 (cross-section reasoning) & \NTtwo & 30 & \LenAccTtwo\% & \LenCITtwo\\
t3 (spec contradiction) & \NTthree & \LenTthree & \LenAccTthree\% & \LenCITthree\\
t4 (spec--RTL consistency) & \NTfour & \LenTfour & \LenAccTfour\% & \LenCITfour\\\midrule
Overall & \NTotal & \LenOverall & \LenAccoverall\% & \LenCIoverall\\
Macro avg.\ (mean of 4 types) & & & \LenMacroAvg\% & \\\bottomrule
\end{tabular}
\caption{Sensitivity analysis: deterministic lenient-threshold scores (t3/t4 location overlap $\ge 0.25$). The conservative Table~\ref{tab:main} is the main result.}
\label{tab:len}
\end{table}

The lenient variant moves t3 from \ConsAccTthree\% to \LenAccTthree\%, t4 from \ConsAccTfour\% to \LenAccTfour\%, and the overall figure from \ConsAccoverall\% to \LenAccoverall\%. As an independent check on the lenient threshold, counting the \AdjudMergedOK\ gray-zone items the adjudication ruled correct on top of the conservative \ConsToverall\ gives \ConsWithGrayOK\ (\ConsAccWithGrayOK\%, 95\% CI \ConsCIWithGrayOK), which sits between the conservative \ConsAccoverall\% and lenient \LenAccoverall\% figures, corroborating that the lenient threshold tracks human-judged correctness. The qualitative pattern is largely unchanged: retrieval and cross-section reasoning stay well above the contradiction-focused types, and the layered location-vs.-category gap persists. One ordering does flip: under conservative scoring t4 (\ConsAccTfour\%) slightly exceeds t3 (\ConsAccTthree\%), while under the lenient threshold t3 (\LenAccTthree\%) exceeds t4 (\LenAccTfour\%), reflecting t3's larger gray zone. The conservative main results are therefore a lower bound on what a more forgiving location rule would credit.

\end{document}